\documentclass[pdflatex,sn-mathphys-num]{sn-jnl}

\usepackage{graphicx}%
\usepackage{multirow}%
\usepackage{amsmath,amssymb,amsfonts}%
\usepackage{amsthm}%
\usepackage{mathrsfs}%
\usepackage[title]{appendix}%

\usepackage{textcomp}%
\usepackage{manyfoot}%
\usepackage{booktabs}%
\usepackage{algorithm}%
\usepackage{algorithmicx}%
\usepackage{algpseudocode}%
\usepackage{listings}%

\usepackage{url}
\usepackage{pifont}
\usepackage{enumitem}
\usepackage{pgfplots}
\pgfplotsset{compat=1.18}
\usepackage{pgfplotstable}
\usepackage{xcolor}
\usepackage{subcaption}
\newcommand{\cmark}{\textcolor{green!60!black}{\ding{51}}}
\newcommand{\xmark}{\textcolor{red}{\ding{55}}}

\theoremstyle{thmstyleone}%
\theoremstyle{thmstyletwo}%

\theoremstyle{thmstylethree}%

\begin{document}

\title[DoorDet: Semi-Automated Multi-Class Door Detection Dataset via Object Detection and Large Language Models]{DoorDet: Semi-Automated Multi-Class Door Detection Dataset via Object Detection and Large Language Models}


\author*[1]{\fnm{Licheng} \sur{Zhang}}\email{licheng.zhang@student.unimelb.edu.au}

\author[1]{\fnm{Bach} \sur{Le}}\email{bach.le@unimelb.edu.au}

\author[1]{\fnm{Naveed} \sur{Akhtar}}\email{naveed.akhtar1@unimelb.edu.au}

\author[1]{\fnm{Tuan} \sur{Ngo}}\email{dtngo@unimelb.edu.au}

\affil[1]{\orgname{The University of Melbourne}, \orgaddress{\street{Parkville}, \city{Melbourne}, \postcode{3010}, \state{Victoria}, \country{Australia}}}


\abstract{Accurate detection and classification of diverse door types in floor plans drawings is critical for multiple applications, such as building compliance checking, and indoor scene understanding. Despite their importance, publicly available datasets specifically designed for fine-grained multi-class door detection remain scarce. In this work, we present a semi-automated pipeline that leverages a state-of-the-art object detector and a large language model (LLM) to construct a multi-class door detection dataset with minimal manual effort. Doors are first detected as a unified category using a deep object detection model. Next, an LLM classifies each detected instance based on its visual and contextual features. Finally, a human-in-the-loop stage ensures high-quality labels and bounding boxes. Our method significantly reduces annotation cost while producing a dataset suitable for benchmarking neural models in floor plan analysis. This work demonstrates the potential of combining deep learning and multimodal reasoning for efficient dataset construction in complex real-world domains.}




\maketitle

\section{Introduction}\label{sec1}
Architectural floor plan drawings are fundamental to the design, analysis, and regulation of building layouts, and they are widely used in both traditional computer-aided design (CAD) and modern architectural workflows \cite{kalervo2019cubicasa5k}. Automatic interpretation of these drawings has attracted increasing interest in the computer vision and machine learning communities due to its broad range of applications, including building information modeling (BIM) \cite{pizarro2022automatic}, building compliance checking (BCC) \cite{chen2024automated}, and automated spatial analysis \cite{zeng2019deep}. Among the key structural elements in floor plans, doors play a critical role, serving not only as physical connections between spaces but also as indicators of accessibility, circulation patterns, and functional zoning within a building \cite{zhang2021holistic}. Table \ref{tab:doordataset} summarizes the presence of door categories across various datasets, showing that the majority include door annotations, which underscores the critical role of doors in floor plan interpretation.

Although most existing floor plan datasets include the door category, they typically treat all doors as a single class. While some datasets \cite{fan2021floorplancad, david2023getting} provide annotations for different door types, these distinctions are largely structural (e.g., single, double, sliding) rather than functional, and thus do not meet the requirements of our task.

Knowing the functional type of each door offers several benefits. Primarily, it aids in building compliance checking and enhances indoor scene understanding. For instance, fire safety is a fundamental aspect of modern architectural design worldwide, aimed at protecting human life and property \cite{nfpa101_2021}. At its core, the provision and placement of emergency exit doors play a pivotal role in emergency route planning, accessibility assessment, and automated building code compliance checking. Additionally, having a multi-class door detection dataset enables us to evaluate different object detection methods and determine whether they work effectively for this specific task and domain. As a result, there is an increasing need for a publicly available dataset with high-quality, fine-grained annotations of diverse door types.

To create an object detection dataset specifically for multi-class door detection, a straightforward approach is to manually annotate each image from scratch. This requires annotators to draw bounding boxes that precisely align with the boundaries of each door and to carefully analyze the surrounding context to determine the appropriate category. However, this manual process is both time-consuming and labor-intensive. To overcome these challenges, in this paper, we propose a novel dataset construction approach that significantly reduces the need for human intervention.

Recent advancements in deep learning-based object detection have yielded impressive results across diverse domains, including remote sensing \cite{feng2024position, wang2024uav}, 3D object detection \cite{xia2024pcdr, zhu2023sfss}, small object detection \cite{wang2024coarse, chai2024mitigate, wei2024review}, salient object detection \cite{yuan2024fgnet, xu2023multi}, etc. Concurrently, large language models (LLMs) or vision-language models (VLMs), such as GPT-4o \cite{hurst2024gpt}, have demonstrated strong capabilities in understanding semantics, context, and structural relationships within complex data, including 3D CAD models \cite{zhang2025largelanguagemodelscomputeraided}. These emerging technologies open new avenues for automatically identifying key architectural components and reasoning about their functional roles.

\begin{table*}[t]
\tiny
\setlength{\tabcolsep}{2.5pt}
    \centering
    \caption{Summary of door category availability in existing floor plan datasets. `BBox' refers to bounding box. `M' is short for million.}
    \begin{tabular}{c|c|c|c|c|c|c|c}
        \toprule
        \textbf{Dataset} & \textbf{Door Availability} & \textbf{BBox Availability} & \textbf{Different Types} & \textbf{Functional} & \textbf{Year} & \textbf{Size} & \textbf{Public}\\
        \midrule
        SESYD \cite{delalandre2010generation} & \cmark & \cmark & \xmark & \xmark & 2010 & 1000 & \cmark \\
        FPLAN-POLY \cite{rusinol2010relational} & \cmark & \xmark & \xmark & \xmark & 2010 & 42 & \cmark \\
        LIFULL HOME'S \cite{nii_lifull2015} & \xmark & \xmark & \xmark & \xmark & 2015 & 5.33 M & \cmark \\
        CVC-FP \cite{de2015cvc} & \cmark & \xmark & \xmark & \xmark & 2015 & 122 & \cmark \\
        Rent3D \cite{liu2018rent3d} & \cmark & \cmark & \xmark & \xmark & 2015 & 215 & \cmark \\
        SydneyHouse \cite{chu2016housecraft} & \cmark & \cmark & \xmark & \xmark & 2016 & 174 & \cmark \\
        R-FP500 \cite{dodge2017parsing} & \cmark & \cmark & \cmark & \xmark & 2017 & 500 & \xmark \\
        Raster-to-Vector \cite{liu2017raster} & \xmark & \xmark & \xmark & \xmark & 2017 & 870 & \cmark \\
        ROBIN \cite{sharma2017daniel} & \cmark & \cmark & \xmark & \xmark & 2017 & 510 & \cmark \\
        RPLAN \cite{wu2019data} & \xmark & \xmark & \xmark & \xmark & 2019 & 80000 & \cmark \\
        BRIDGE \cite{goyal2019bridge} & \cmark & \cmark & \xmark & \xmark & 2019 & 13000 & \cmark \\
        R2V \cite{zeng2019deep} & \cmark & \xmark & \xmark & \xmark & 2019 & 815 & \cmark \\
        R3D \cite{zeng2019deep} & \cmark & \xmark & \xmark & \xmark & 2019 & 232 & \cmark \\
        CubiCasa5K \cite{kalervo2019cubicasa5k} & \cmark & \xmark & \xmark & \xmark & 2019 & 5000 & \cmark \\
        SCUT-AutoALP \cite{liu2020scut} & \xmark & \xmark & \xmark & \xmark & 2020 & 602 & \cmark \\
        Rent3D++ \cite{Vidanapathirana_2021_CVPR} & \cmark & \xmark & \xmark & \xmark & 2021 & 215 & \cmark \\
        ZSCVFP \cite{dong2021vectorization} & \cmark & \xmark & \xmark & \xmark & 2021 & 10800 & \xmark \\
        RuralHomeData \cite{lu2021data} & \cmark & \cmark & \xmark & \xmark & 2021 & 800 & \xmark \\
        RFP \cite{Lv_2021_CVPR} & \cmark & \cmark & \xmark & \xmark & 2021 & 7000 & \xmark \\
        Kim et al. \cite{kim2021automatic} & \cmark & \xmark & \xmark & \xmark & 2021 & 230 & \xmark \\
        RUB \cite{simonsen2021generalizing} & \cmark & \xmark & \xmark & \xmark & 2021 & 303 & \cmark \\
        FloorPlanCAD \cite{fan2021floorplancad} & \cmark & \cmark & \cmark & \xmark & 2021 & 15663 & \cmark \\
        David et al. \cite{david2023getting} & \cmark & \cmark & \cmark & \xmark & 2023 & 35000 & \xmark \\
        MLSTRUCT-FP \cite{pizarro2023large} & \xmark & \xmark & \xmark & \xmark & 2023 & 954 & \cmark \\
        Park et al. \cite{park2024developing} & \cmark & \xmark & \xmark & \xmark & 2024 & 10000 & \xmark \\
        MSD \cite{van2024msd} & \cmark & \xmark & \xmark & \xmark & 2024 & 5372 & \cmark \\
        ArchCAD-400K \cite{luo2025archcad} & \cmark & \cmark & \xmark & \xmark & 2025 & 413062 & \cmark \\
        \textbf{DoorDet} (\textbf{Ours}) & \cmark & \cmark & \cmark & \cmark & 2025 & 4991 & \cmark \\
        \bottomrule
    \end{tabular}
    \label{tab:doordataset}
\end{table*}

Specifically, we leverage a state-of-the-art object detector to ensure high-precision door localization. In addition, a state-of-the-art LLM is employed to identify different door types in architectural floor plans. Our system is designed to detect all door instances, and subsequently distinguish them among various types. This prediction process is informed not only by the visual characteristics of door symbols but also by contextual cues, such as their spatial connectivity to rooms and the presence of relevant text annotations. Afterward, a human-in-the-loop refinement process is introduced, enabling domain experts to review and correct the automated predictions.

As a key outcome of our work, we have successfully constructed a curated dataset comprising real-world architectural floor plans annotated with detailed door type labels. Unlike general architectural datasets that focus on broader elements such as rooms and walls, our dataset emphasizes fine-grained classification of door instances based on their functions. We posit that this dataset will play a crucial role in advancing research in building code compliance automation, indoor semantic analysis, and object detection. The dataset will be made publicly available upon acceptance.

To sum up, our contributions include the following.
\begin{itemize}
\item We propose a novel methodology for object detection dataset construction, which we argue generalizes well to a broad range of related tasks. Compared to conventional dataset generation approaches, our method significantly reduces the need for manual effort.
\item We construct a curated dataset, \textbf{DoorDet}, consisting of annotated door instances derived from real-world floor plans. The dataset features fine-grained labels that distinguish multiple door categories. To the best of our knowledge, this is the first object detection dataset to include detailed door types categorized by functional roles.
\item We benchmark \textbf{DoorDet} using multiple state-of-the-art object detection models, demonstrating its utility for both training and evaluation in object detection tasks.
\item We demonstrate the cross-domain generalizability of models trained on \textbf{DoorDet}, showing effective adaptation across various architectural layout datasets.
\end{itemize}
\section{Related Work}\label{sec2}
\subsection{Object Detection in Floor Plans}
Object detection in floor plans is an active research direction. Xu et al. \cite{xu2024multiscale} employed a multi-scale detection strategy with multiple detection heads to enhance the detection of small and overlapping objects in floor plans. Likewise, Shehzadi et al. \cite{shehzadi2022mask} introduced a semi-supervised learning framework based on Mask R-CNN \cite{he2017mask} enhanced with a student-teacher architecture, which effectively learnt from a small fraction of labeled data supplemented with a larger set of unlabeled data. Similarly, Mishra et al.~\cite{mishra2021towards} utilized Cascade Mask R-CNN \cite{cai2018cascade}, integrating deformable convolution to better capture geometric variations of objects in floor plans. Recently, Jakubik et al.~\cite{jakubik2022designing} have identified uncertain symbols in floor plans and actively queried human annotators for clarification. Building upon earlier efforts, Lv et al.~\cite{Lv_2021_CVPR} employed yolov4 \cite{bochkovskiy2020yolov4} with Mosaic data augmentation to detect and group text and symbols on floor plans, improving semantic context and accuracy in identifying room types. In another related approach, Lu et al. \cite{lu2021data} simultaneously recognized graphical elements and detected room-type texts, addressing the intertwined nature of structural and semantic information in floor plans.

Additionally, in \cite{park20213dplannet}, the object detection component of 3DPlanNet was designed to identify architectural elements such as walls, doors, windows, and rooms within 2D floor plan images. Other efforts, such as those by Surikov et al. \cite{surikov2020floor}, detected architectural elements such as doors, windows, and furniture using Faster R-CNN \cite{ren2016faster}. In a related study, Sch{\"o}nfelder et al.~\cite{schonfelder2024deep} focused on detecting textual regions in architectural floor plans and evaluated multiple object detection methods: yolov5 \cite{yolov5}, yolov7 \cite{wang2023yolov7}, yolov8 \cite{yolov8}, yolor \cite{wang2023you}, and Faster R-CNN \cite{ren2016faster}. In another line of work, Oh et al. \cite{oh2025integrating} analyzed construction site images and architectural drawings to identify and localize elements related to finishing works, such as masonry and tiling, using yolov5 \cite{yolov5}. Another relevant contribution was by Park et al. \cite{park2024developing}, who utilized yolov3 \cite{redmon2018yolov3} for detecting architectural objects in floor plan images, achieving high accuracy by training on a large, richly annotated dataset of spatial layouts.
\subsection{Datasets for Floor Plan Analysis}
Delalandre et al. \cite{delalandre2010generation} developed the SESYD dataset, a synthetic collection for benchmarking symbol recognition and spotting systems. Similarly, Rusi{\~n}ol et al. \cite{rusinol2010relational} gathered 42 high-resolution architectural floor plan images from four architectural projects, annotated with 38 symbol classes. Further, the dataset in \cite{de2015cvc} consisted of 122 scanned floor plan documents with ground-truth annotations for structural symbols like rooms, walls, doors, windows, parking doors, and separations. In another line of work, in \cite{liu2018rent3d}, the Rent3D dataset consisted of 1570 images captured from 215 apartments, each annotated with detailed floor-plan information including walls, doors, and windows. It provided ground truth 3D layout data and floor-plan priors to facilitate research on monocular indoor layout estimation and scene understanding. Extending this effort, Chu et al. \cite{chu2016housecraft} introduced the SydneyHouse dataset, comprising 174 residential buildings annotated with floor plan alignment, 3D structure, and textures, where 3D annotations included height, window/door locations, and pose. In a related contribution, Dodge et al. \cite{dodge2017parsing} presented the R-FP500 dataset with 500 floor plan images collected from real estate websites, manually annotated for walls, architectural objects, and text labels. Additionally, Liu et al. \cite{liu2017raster} annotated 870 floor plan images with walls, doors, icons, and junctions. Meanwhile, Sharma et al. \cite{sharma2017daniel} created the ROBIN dataset containing 510 real floor plan images with detailed room and furniture symbol annotations.

On a larger scale, Wu et al. \cite{wu2019data} collected real residential floor plans from a Chinese real estate website and automatically parsed and annotated room locations, shapes, and types using a combination of heuristic rules and pre-trained models, resulting in over 80000 floor plans across 21 room types. Similarly, Goyal et al. \cite{goyal2019bridge} introduced the BRIDGE dataset, sourced from architectural websites and public sources, annotated with bounding boxes and labels for common architectural elements like doors, windows, and furniture. Along the same line, Zeng et al. \cite{zeng2019deep} prepared two datasets: R2V and R3D, which included floor plan images with annotations of room boundaries and types, supplemented by additional floor plan images. Moreover, Kalervo et al. \cite{kalervo2019cubicasa5k} annotated 5000 floor plan images with polygon-level labels for rooms, walls, doors, windows, furniture, and fixtures, covering over 80 object classes.

Continuing this trend, Liu et al. \cite{liu2020scut} developed the SCUT-AutoALP dataset containing 602 annotated samples, divided into residential floor plans (300 images with layout, boundary, and attribute labels) and urban campus plans (302 images with the same annotations). To extend prior datasets, Vidanapathirana et al. \cite{Vidanapathirana_2021_CVPR} introduced Rent3D++, an augmented dataset extending Rent3D \cite{liu2018rent3d}, which included 215 floor plans from rental listings, 1570 room photographs, annotated with room types, walls, doors, windows, and real-world scale and layout information. Similarly, Dong et al. \cite{dong2021vectorization} developed a private dataset with 10800 color floor plan images encompassing a variety of decorative textures and styles. In parallel, Lu et al. \cite{lu2021data} constructed a dataset of 800 real-world floor plan images from Chinese rural residences annotated with geometric components such as walls, doors and windows, as well as semantic labels including room types like bedrooms, kitchens, and storage rooms, and associated textual annotations extracted from the plans.

Also, Lv et al. \cite{Lv_2021_CVPR} introduced a dataset with around 7000 annotated instances of rooms, doors, walls, windows, and furniture from residential floor plans, featuring detailed labels of room types and architectural elements to support floor plan recognition and reconstruction. Further, the dataset in \cite{kim2021automatic} featured floor plan images of large-scale complex buildings, where a scale matching module handled varying scales across floor plans, and each plan was divided into uniform patches labeled with object classes such as walls, doors, windows, elevators, and stairwells. From another perspective, Simonsen et al. \cite{simonsen2021generalizing} converted CAD files into graphs with nodes representing geometric primitives and edges encoding their topological and geometric relationships. In addition, the FloorPlanCAD dataset \cite{fan2021floorplancad} included over 15000 CAD floor plans preserving precise geometric and semantic information, with fine-grained annotations for 30 object categories such as stairs, furniture, doors, windows, and walls. Although the dataset included door locations and distinguished between basic door types, such as single, double, and sliding doors, these categories were simple and did not convey the functional role of the doors.

More recently, David et al. \cite{david2023getting} generated a dataset of 35000 annotated door images extracted from CubiCasa5K \cite{kalervo2019cubicasa5k}, focusing specifically on door identification and classification, alongside a literature review of key floor plan datasets. The dataset provided both door positions and types but the classification was fairly simple, concentrating on hinge side including left or right, opening direction like inward outward or swinging, and door style such as single, double, sliding or folding. Similarly, in \cite{pizarro2023large}, the MLSTRUCT-FP dataset included 954 floor plan images used for training and evaluating wall segmentation models, with over 1.3 million image patches extracted to capture detailed structural features. In a large-scale web mining effort, Park et al. \cite{park2024developing} collected a large dataset of 100000 data rows through extensive web crawling, categorized into 21 classes and three spatial relationship categories. Furthermore, in \cite{van2024msd}, van Engelenburg et al. introduced the MSD dataset, comprising 5372 floor plans of medium to large-scale building complexes, encompassing over 18900 distinct apartments. Each floor plan was available in image, vectorized, and graph-based formats. Lastly, ArchCAD-400K \cite{luo2025archcad} was a large-scale, publicly available dataset containing over 413000 annotated chunks from 5538 architectural CAD floor plans. Leveraging CAD layers and blocks, it automated detailed annotations across 27 categories including doors and structural elements, covering mostly commercial and public buildings. This dataset provided a valuable benchmark for panoptic symbol spotting and architectural CAD analysis.
\section{Methodology}\label{sec3}
\begin{figure*}[t]
    \centering
    \includegraphics[width=1.0\linewidth]{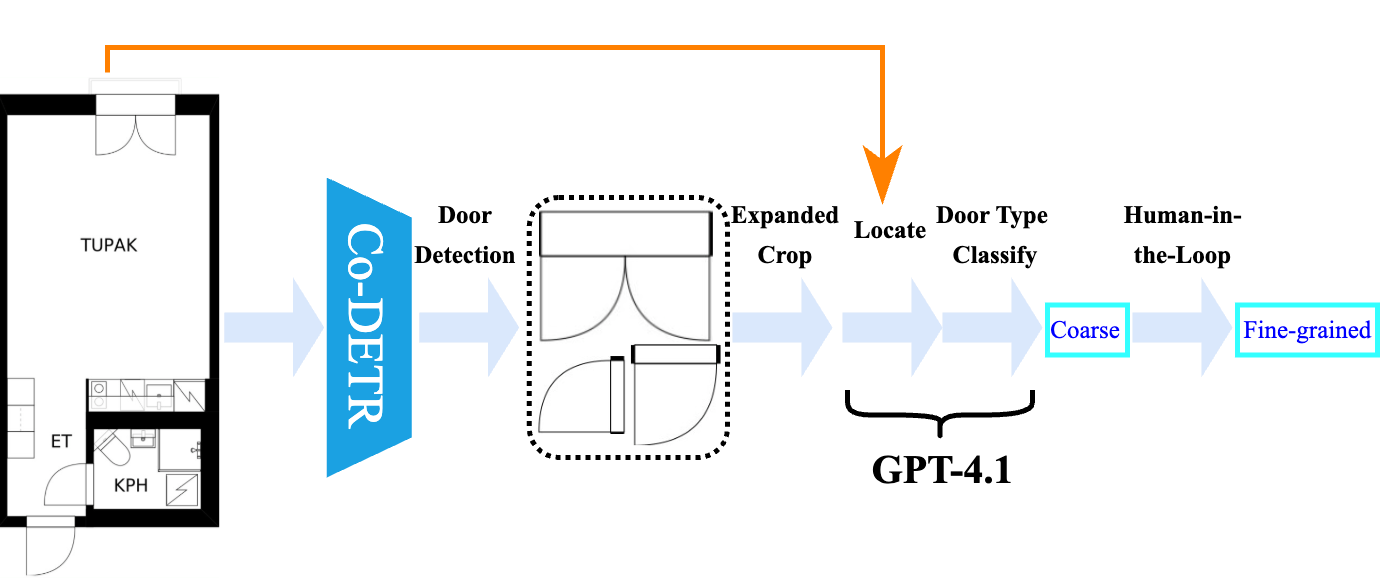}
    \caption{Our proposed vision-language-driven human-in-the-loop pipeline for multi-class door detection dataset generation.
    \label{fig:overview}}
\end{figure*}

In this paper, we propose a novel human-in-the-loop-assisted, vision-language-driven approach for constructing a multi-class door detection dataset, as illustrated in Figure~\ref{fig:overview}. The pipeline begins with a state-of-the-art object detector that identifies all door instances as a single class. Subsequently, heuristic techniques are applied to facilitate downstream processing. A state-of-the-art LLM is then employed to both locate and predict the type of individual door instances. Finally, a human-in-the-loop refinement stage is introduced to transform the initial coarse annotations into high-quality, fine-grained labels. Our approach enables the efficient construction of a practically valuable dataset with significantly reduced manual effort and time cost, compared to conventional dataset creation methods.
\subsection{Single-Class Door Detection}\label{sec3.1}
Benefiting from existing publicly available single-class door detection datasets, we can effortlessly train a unified model that detects all door types as a single class, applicable across diverse real-world CAD drawings. To ensure high accuracy in door detection on out-of-domain data, we adopt a state-of-the-art object detector that achieves top-tier performance on the COCO benchmark \cite{lin2014microsoft}, which is Co-DETR \cite{zong2023detrs}. Co-DETR exhibits strong performance in detecting small and dense objects, making it particularly suitable for floor plan scenarios where door symbols are often compact and tightly clustered. Accordingly, we adopt Co-DETR as our object detector. Compared to its predecessor DETR \cite{carion2020end}, which suffers from slow convergence and limited performance on dense or small object detection due to its merely one-to-one label assignment strategy, Co-DETR introduces a collaborative hybrid assignment mechanism, which combines one-to-one and many-to-one assignments during training, significantly improving both convergence speed and detection accuracy. In Co-DETR, the primary decoder performs one-to-one matching (as in DETR), used during both training and inference. In parallel, the auxiliary decoder adopts many-to-one matching (as in Faster R-CNN), used only during training and discarded at inference time. Both decoders share a common backbone and encoder but differ in their attention weights and matching strategies. Compared with DETR, Co-DETR achieves up to 2$\times$ faster convergence and improved performance on challenging detection tasks.

Mathematically, given an input image $\mathbf{I}$, the detector outputs a set of $N$ predicted door instances:
\begin{equation}
    \{(b_i, c_i)\}_{i=1}^N = \text{Co-DETR}(\mathbf{I}),
\end{equation}
where $b_i$ and $c_i$ denote the bounding box and confidence score of door $i$, respectively. The detection network is trained to detect only a single object category, in this case doors, by distinguishing between background and door instances.
\subsection{Door Type Prediction via LLMs}\label{sec3.2}
\subsubsection{Motivation}
Although object detectors effectively localize doors, they generally lack the semantic and contextual reasoning capabilities required to distinguish between functional door types. To address this limitation, we propose leveraging a vision-capable LLM to infer door functionality directly from visual and contextual cues. Recent advancements in LLMs with vision capabilities, such as GPT-4o \cite{hurst2024gpt}, enable the interpretation of images, diagrams, and floor plans in conjunction with textual information \cite{zentgraf2024enhancing}. Their ability to integrate visual and semantic context makes them well-suited for high-level interpretation of functional elements within architectural layouts. GPT-4.1 \cite{openai2025gpt41}, one of OpenAI’s latest flagship models, supports multimodal inputs and introduces a one-million-token context window, significantly enhancing its capacity to reason over complex documents and large-scale visual layouts. Consequently, we adopt GPT-4.1 as the LLM in this work.
\subsubsection{Door Type Prediction}
After obtaining the generic door detection model described in Section \ref{sec3.1}, we first perform inference on our selected floor plan images using the optimized model. This phase locates all door instances within each image. Subsequently, we process each detected door individually and assign a specific type to it.

For each detected door instance, although we have its position within the image, LLMs are not well-suited to directly infer coordinates from positional values. Therefore, we provide the LLM with both the cropped door region and the full floor plan image, and then ask it to locate the corresponding region within the full floor plan.

However, during implementation, we observed that providing only the cropped door region makes it difficult for the LLM to accurately identify the specific door, due to the similar appearance of doors and the limited contextual cues. To address this, we expand the cropped region by extending the bounding box with a fixed margin $m$ pixels on all sides (e.g., $m$ = 100) to include more surrounding context. Additionally, since the expanded region may contain multiple doors, we explicitly instruct the LLM to focus only on the central door within the cropped image.

Formally, door localization via LLM can be expressed as:
\begin{equation}
L_i = \text{LLM}(C_i, \mathbf{I}),
\end{equation}
where $C_i$ is the expanded cropped image patch of the $i$-th detected door extracted from the full floor plan $\mathbf{I}$, and $L_i$ denotes the estimated location information of the door (e.g., room reference or spatial description within $\mathbf{I}$).

After locating the door within the complete floor plan, we then prompt the LLM to identify the door type by analyzing its connectivity to surrounding rooms and leveraging semantic and contextual cues present in the layout.

Analytically, the door type prediction can be formulated as:
\begin{equation}
    T_i = \mathrm{LLM}\big(L_i, \mathbf{I} \mid \mathcal{P} \big),
\end{equation}
where $L_i$ denotes the location of the $i$-th detected door. $\mathbf{I}$ denotes the full floor plan image. $\mathcal{P}$ denotes the prompt including the task instruction, the door category definitions, as well as the output format. $T_i$ denotes the predicted door category. The LLM integrates all these information and outputs the door type according to the rules outlined in $\mathcal{P}$. The output format is a category number or a combination (e.g., emergency exit combined with main entry) as specified.
\subsection{Human-in-the-Loop Refinement}\label{sec3.3}
Inevitably, LLM predictions cannot achieve perfect accuracy, and some degree of error is expected within the resulting dataset. To address this, we introduce a human-in-the-loop refinement stage, where trained annotators systematically verify and refine the coarse annotations produced by the model. We instruct annotators to correct mislabeled instances, add missing detections, and remove false positives. Additionally, they adjust bounding box positions when the original boxes fail to fully enclose the door or include excessive non-door regions. This refinement process is substantially faster than annotating from scratch, which not only requires drawing bounding boxes but also demands careful consideration of the correct category for each instance, a task that is both time-consuming and labor-intensive. By starting with model-generated predictions and applying targeted corrections, we greatly reduce the manual workload while still ensuring high-quality annotations.

To assess the impact of introducing the human-in-the-loop mechanism, we analyze the correlation between task difficulty and the presence of human intervention within our proposed framework for multi-class door detection.

\paragraph{Task Formulation}
Let $\mathcal{T}$ denote the set of detection tasks, where each task corresponds to detecting a specific door type. For each task $t \in \mathcal{T}$, we define $D(t) \in \mathbb{R}_{\ge 0}$ as the \emph{task difficulty}, and $\Delta(t) \in \mathbb{R}$ as the \emph{performance gain} resulting from human intervention. 

Our goal targets at quantifying the correlation between $D(t)$ and $\Delta(t)$: $\operatorname{Corr}_{t \in \mathcal{T}} \bigl( D(t), \Delta(t) \bigr)$. A larger positive value indicates a stronger correlation between them.

We define the task difficulty $D(t)$ as the initial detection error obtained by the model before human intervention:
\begin{equation}
D(t) = 1 - \text{Accuracy}_{\text{model-only}}(t),
\label{eq:difficulty}
\end{equation}
where `model-only' indicates only relying on models, such as object detector and LLM, while without human-in-the-loop refinement. The lower the accuracy gets, the harder the task becomes.

The performance gain $\Delta(t)$ is measured as the improvement in accuracy resulting from human intervention:
\begin{equation}
\Delta(t) = \text{Accuracy}_{\text{HITL}}(t) - \text{Accuracy}_{\text{model-only}}(t),
\label{eq:gain}
\end{equation}
where `HITL' denotes including the human-in-the-loop setup. A larger value of 
$\Delta(t)$ indicates a greater contribution of human feedback to performance improvement.

The above formulations enable us to analyze how the effectiveness of human feedback correlates with the inherent difficulty of each task in our multi-class door detection framework.
\section{DoorDet: Proposed Multi-Class Door Detection Dataset}\label{sec4}
Our dataset, \textbf{DoorDet}, is constructed from the CubiCasa5K dataset \cite{kalervo2019cubicasa5k}, which provides a diverse collection of floor plans encompassing a wide range of architectural styles and spatial layouts. The dataset consists of 5000 samples from real residential buildings across multiple countries, divided into 4200 training, 400 validation, and 400 test instances. These samples represent multi-room and multi-floor structures, including both apartments and standalone houses. Many floor plans include textual annotations indicating room types such as bedrooms, bathrooms, and kitchens, along with standardized architectural symbols for doors and other structural elements. The images feature minimal decorative content and maintain consistent symbology. The embedded text within the plans is particularly valuable for inferring room functions, which can in turn facilitate the prediction of corresponding door types.

We first extract floor plan images from the source SVG files in the CubiCasa5K dataset, rather than relying on the pre-rendered raster images provided, and retain the native scale and precision of the floor plans. The annotations are then generated using our proposed method described in Section~\ref{sec3}, which automatically produces coarse annotations via Co-DETR and GPT-4.1. These are subsequently refined through a human-in-the-loop process that corrects mispredictions, removes false positives, and results in high-quality, fine-grained annotations for each door instance.

For the first step, which is to train a single-class door detection model, we construct a larger dataset by combining two publicly available single-class door detection datasets, which contain 1047 and 837 training examples, respectively \cite{doorobjectdetection, floorplans500}.

Our constructed \textbf{DoorDet} dataset includes a variety of door types which are detailed as follows.
\begin{itemize}[noitemsep, topsep=0pt]
\item \textbf{Main entry door:} This refers to the primary entrance to a building or unit, typically connecting interior spaces to the exterior or shared access areas.
\item \textbf{Bedroom door:} This type of door leads to sleeping areas within the interior of the building.
\item \textbf{Bathroom or washroom door:} These doors provide access to hygiene-related spaces such as toilets, bathrooms, or showers.
\item \textbf{Kitchen door:} A kitchen door connects the kitchen to adjacent spaces such as dining rooms, hallways, or utility areas.
\item \textbf{Living room or dining room door:} These doors lead to communal living spaces such as lounges, living rooms, or dining areas.
\item \textbf{Laundry or utility room door:} This type of door provides access to service areas such as laundry rooms, mechanical rooms, or utility closets. Any door that does not fit clearly into another category is typically assigned to this type.
\item \textbf{Garage door:} This is a large door that provides vehicle access to a garage, typically from the outside. Interior doors that allow pedestrian access between the garage and living spaces are also included in this category.
\item \textbf{Balcony or terrace door:} These are external doors that open to outdoor areas such as balconies, terraces, or verandas.
\item \textbf{Emergency exit door:} This door serves as a designated exit during emergencies and is essential for safety compliance and evacuation planning.
\item \textbf{Study room door:} This door provides access to private or semi-private workspaces such as studies, home offices, or libraries. The type is manually determined by referring to example patterns found during the refinement process.
\end{itemize}
\subsection{Data Statistics}
To better understand the scope and richness of our dataset, we summarize its core characteristics below. Table~\ref{tab:dataset-stats} presents the key statistics, including the total number of floor plan images, the average number of doors per image, the average image resolution, and other important attributes. Each image contains multiple door instances spanning a diverse set of categories. Notably, the high image resolution sets \textbf{DoorDet} apart from most existing object detection benchmarks \cite{lin2014microsoft, everingham2010pascal}. Each door instance is annotated with both a category label and a bounding box. To the best of our knowledge, \textbf{DoorDet} is the first dataset in the architectural domain to provide such detailed annotations for door instances, encompassing both their locations and functional types.
\begin{table*}[t]
\centering
\caption{Basic information about the \textbf{DoorDet} dataset. `BBox' refers to bounding box.}
\begin{tabular}{l|c}
\toprule
\textbf{Property} & \textbf{Statistic} \\
\midrule
Total floor plan images      & 4991 \\
Number of door categories    & 10 \\
Average doors per image      & 7.81 \\
Average image resolution     & 1495.2$\times$1310.6 \\
Annotation format            & BBox + type \\
Source                       & CubiCasa5K \cite{kalervo2019cubicasa5k} \\
Train/Val/Test split     & 4192/400/399 \\
\bottomrule
\end{tabular}
\label{tab:dataset-stats}
\end{table*}

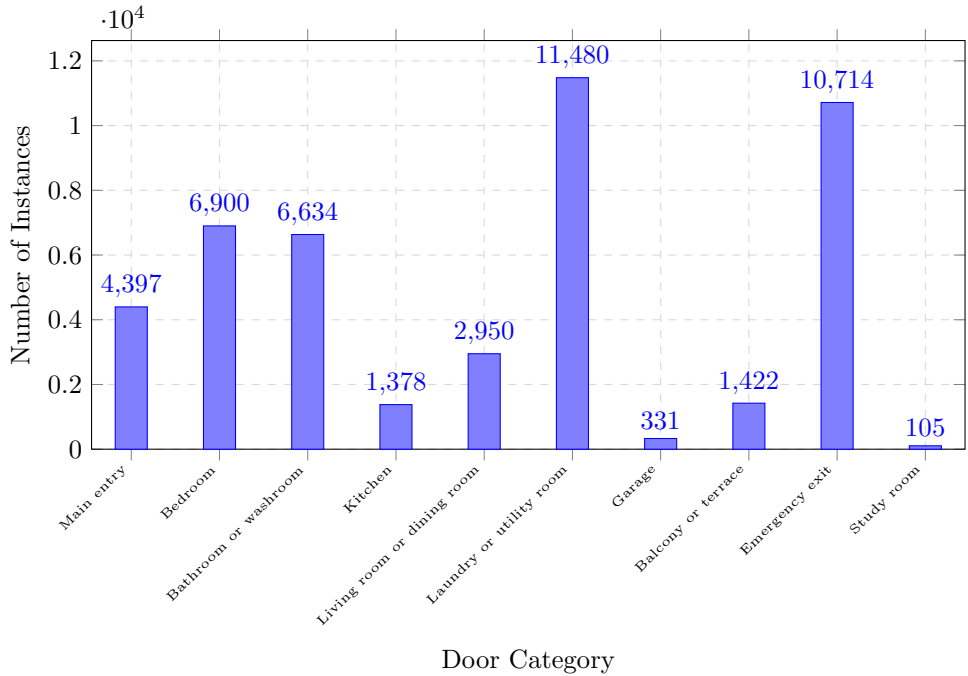
\begin{figure*}[t]
\centering
\begin{tikzpicture}
\begin{axis}[
    ybar,
    bar width=12pt,
    width=\linewidth,
    height=7cm,
    enlarge x limits=0.05,
    ylabel={Number of Instances},
    xlabel={Door Category},
    symbolic x coords={
        Main entry, Bedroom, Bathroom or washroom, Kitchen, Living room or dining room, Laundry or utility room, Garage, Balcony or terrace, Emergency exit, Study room
    },
    xtick=data,
    xticklabel style={rotate=45, anchor=east,font=\tiny},
    nodes near coords,
    nodes near coords align={vertical},
    ymin=0,
    grid=major,
    major grid style={dashed, gray!30}
]
\addplot+[style={fill=blue!50}] coordinates {
    (Main entry,4397)
    (Bedroom,6900)
    (Bathroom or washroom,6634)
    (Kitchen,1378)
    (Living room or dining room,2950)
    (Laundry or utility room,11480)
    (Garage,331)
    (Balcony or terrace,1422)
    (Emergency exit,10714)
    (Study room,105)
};
\end{axis}
\end{tikzpicture}
\caption{Distribution of instances across the 10 door categories in the \textbf{DoorDet} dataset.}
\label{fig:cat-barplot}
\end{figure*}

In addition to these global statistics, we also present the total number of door instances for each category in the dataset. Figure~\ref{fig:cat-barplot} illustrates the distribution of different door categories. As shown in the figure, the study room door and garage door categories contain significantly fewer instances compared to others, while all remaining categories have over 1000 instances, with laundry or utility room doors and emergency exit doors exceeding 10000 instances. This indicates a clear class imbalance within the dataset. Such imbalance is uncommon in conventional object detection benchmarks, making \textbf{DoorDet} a more challenging and realistic testbed for evaluating model robustness in imbalanced scenarios.

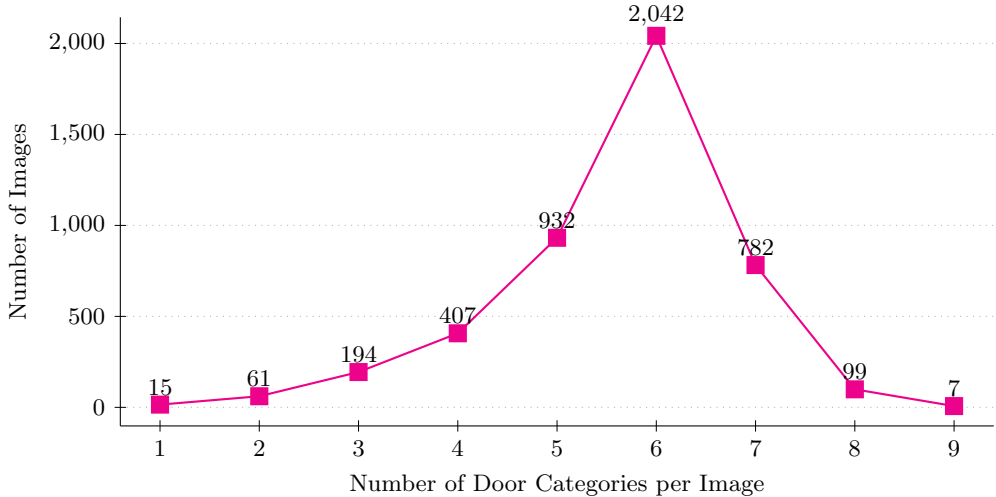
\begin{figure*}[t]
\centering
\begin{tikzpicture}
\begin{axis}[
    width=\linewidth,
    height=7cm,
    axis lines=left,
    axis line style={-},
    xlabel={Number of Door Categories per Image},
    ylabel={Number of Images},
    xtick={1,2,3,4,5,6,7,8,9},
    xticklabels={1,2,3,4,5,6,7,8,9},
    ymin=0,
    ymajorgrids=true,
    grid style={dotted, gray!60},
    tick label style={font=\small},
    label style={font=\small},
    mark size=3pt,
    nodes near coords,
    nodes near coords style={font=\small, anchor=south},
    every node near coord/.append style={black},
    enlargelimits=0.05,
    tick style={black},
    tick label style={font=\small}
]

\addplot[
    color=magenta,
    mark=square*,
    thick
] coordinates {
    (1,15)
    (2,61)
    (3,194)
    (4,407)
    (5,932)
    (6,2042)
    (7,782)
    (8,99)
    (9,7)
};

\end{axis}
\end{tikzpicture}
\caption{Distribution of door categories per image in the \textbf{DoorDet} dataset.}
\label{fig:cat-per-image}
\end{figure*}

Furthermore, we analyze the distribution of the number of door categories per image in Figure~\ref{fig:cat-per-image}. As shown in the figure, most images contain around six types of door instances, which aligns with the nature of floor plan layouts, typically composed of diverse room types, each associated with function-specific doors, indicating the complexity and challenge of accurate door detection and fine-grained classification.

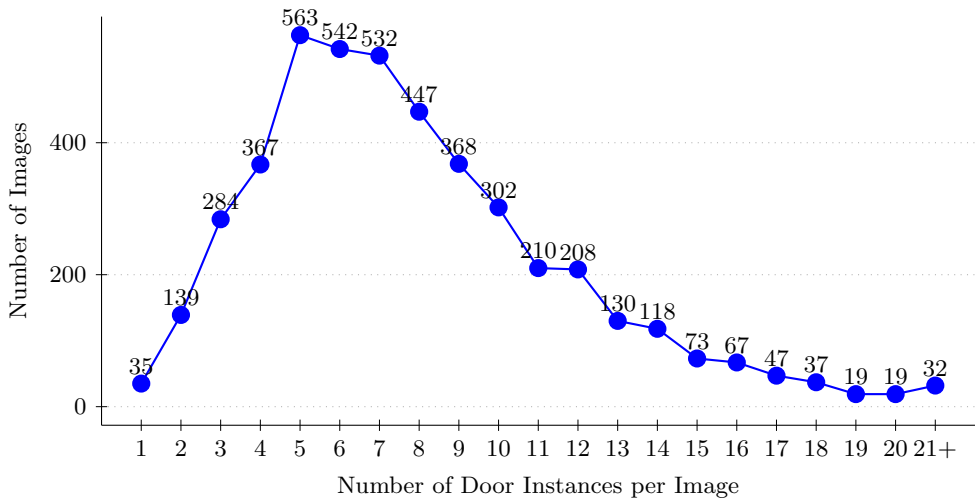
\begin{figure*}[t]
\centering
\begin{tikzpicture}
\begin{axis}[
    width=\linewidth,
    height=7cm,
    axis lines=left,
    axis line style={-},
    xlabel={Number of Door Instances per Image},
    ylabel={Number of Images},
    symbolic x coords={1,2,3,4,5,6,7,8,9,10,11,12,13,14,15,16,17,18,19,20,21+},
    xtick=data,
    ymin=0,
    ymajorgrids=true,
    grid style={dotted, gray!60},
    tick label style={font=\small},
    label style={font=\small},
    mark size=3pt,
    nodes near coords,
    nodes near coords style={font=\small, anchor=south},
    every node near coord/.append style={black},
    enlargelimits=0.05,
    tick style={black}
]

\addplot[
    color=blue, 
    mark=*,
    thick
] coordinates {
    (1,35)
    (2,139)
    (3,284)
    (4,367)
    (5,563)
    (6,542)
    (7,532)
    (8,447)
    (9,368)
    (10,302)
    (11,210)
    (12,208)
    (13,130)
    (14,118)
    (15,73)
    (16,67)
    (17,47)
    (18,37)
    (19,19)
    (20,19)
    (21+,32)
};

\end{axis}
\end{tikzpicture}
\caption{Distribution of the door instances per image in the \textbf{DoorDet} dataset.}
\label{fig:instance-per-image}
\end{figure*}

Finally, we analyze the distribution of the number of door instances per image. Figure~\ref{fig:instance-per-image} shows how many images contain different numbers of annotated door instances. As observed, most images contain more than five doors, which highlights the practical significance and inherent difficulty of the door detection task in the dataset.

\section{Experiments}\label{sec5}
\subsection{Datasets}
As mentioned earlier, we use a combined dataset from \cite{doorobjectdetection, floorplans500} to train a unified single-class door detection model, with training performed on the merged training sets and evaluation on the merged test sets.

Benchmark experiments are conducted on our constructed \textbf{DoorDet} dataset. We follow the original split settings from CubiCasa5K \cite{kalervo2019cubicasa5k}, using its training, validation, and test file lists to define the corresponding splits in our dataset.
\subsection{Benchmarking Methods}
\textbf{Co-DETR \cite{zong2023detrs}:} For Co-DETR, we adopt the best-performing Co-DETR variant with a Vision Transformer Large (ViT-L) \cite{dosovitskiy2020image} backbone. It achieves a 66.0 mAP on the COCO \texttt{test-dev}. We utilize 300 learned object queries for it. The model is optimized using the AdamW optimizer with an initial learning rate of $1 \times 10^{-4}$, which is decayed by a factor of 0.5 at epochs 5, 6, 7, and 10, respectively, and then kept fixed for the remainder of training, up to around 20 epochs, which is stopped once accuracy plateaus. Batch size is set to 2. The experiments are performed using two NVIDIA A100 GPUs. The network is initialized with the pretrained weights of Co-DETR on the COCO dataset \cite{lin2014microsoft}. During inference, the object confidence threshold is set to 0.3.

\textbf{InternImage-H \cite{wang2023internimage}:} For InternImage, we adopt the InternImage-H backbone, which contains approximately 1 billion parameters and incorporates the novel DCNv3 deformable convolution operator. It achieves a 65.5 mAP on the COCO \texttt{test-dev}. The detection framework is based on DINO \cite{zhangdino}, and we apply multi-scale training to enhance performance. Optimization is performed using the AdamW optimizer with a starting learning rate of $1 \times 10^{-4}$, combined with layer-wise learning rate decay and a weight decay of $1 \times 10^{-4}$. Training is conducted for 45000 iterations, including a 500-iteration warm-up phase. The learning rate is decayed by a factor of 0.1 at 20000 and 40000 iterations, respectively. The model is trained with a batch size of 1 using one NVIDIA A100 GPU. The network is initialized with pretrained InternImage-H weight on the COCO dataset \cite{lin2014microsoft}.

\textbf{Focal-Stable-DINO \cite{ren2023strong}:} Focal-Stable-DINO combines the FocalNet-Huge backbone \cite{yang2022focal} with the Stable-DINO detection head \cite{liu2023detection}, achieving 64.8 mAP on COCO \texttt{test-dev} without any test-time augmentation. FocalNet-Huge provides strong feature representation while maintaining computational efficiency. Stable-DINO introduces a position-supervised classification loss and a position-modulated matching cost, which together stabilize decoder-layer predictions and eliminate the unstable matching paths commonly found in standard DETR variants. The model is optimized using AdamW with a base learning rate of $1 \times 10^{-4}$ and a backbone-specific learning rate scaled by a factor of 0.1. The weight decay is set to $1 \times 10^{-4}$. Training is conducted on two NVIDIA A100 GPUs with a batch size of 2, for a total of 120000 iterations. The learning rate is decayed by a factor of 0.1 at 82500 and 110000 iterations, respectively. The number of de-noising queries is set to 1000. The parameters of FocalNet-Huge are initialized with pretrained weights on ImageNet \cite{deng2009imagenet}, obtained via the timm library \cite{rw2019timm}.

\textbf{EVA \cite{fang2023eva}:} We employ an EVA-style Vision Transformer Giant (ViT-G) \cite{zhai2022scaling} backbone featuring 40 Transformer layers, each with 16 attention heads and an embedding dimension of 1408. The model processes inputs at a resolution of 1280$\times$1280 with patch size 16 and uses windowed attention with windows of size 16, interleaved with global attention every four blocks. Regularization includes a drop path rate of 0.6 and activation checkpointing to reduce memory usage. It achieves a 64.7 mAP on COCO \texttt{test-dev}. Training uses the AdamW optimizer with a base learning rate of $2.5 \times 10^{-5}$, decayed layer-wise by a factor of 0.9 across the 40 layers. The learning rate is set to $1 \times 10^{-5}$ at iteration 35000 and further reduced to $1 \times 10^{-6}$ at iteration 40000, over a total of 45000 training steps, with a warmup of 500 iterations at the beginning. The batch size is set to 8. Training is conducted using two NVIDIA A100 GPUs. We initialize the model using weights pretrained on the COCO dataset \cite{lin2014microsoft}.
\subsection{Implementation details}
For GPT-4.1, we access its capabilities through the OpenAI API. $m$ is set to 200.
\subsection{Evaluation Metrics}
We adopt mAP (mean Average Precision at IoU thresholds from 0.50 to 0.95), and mAP@50 (mean Average Precision at IoU = 0.50) to assess the performance of single-class and multi-class door detection. Since the bounding boxes in our dataset do not precisely align with the exact boundaries of door instances, we additionally use mAP@50 as the evaluation metric to provide a more robust and tolerant measure of detection performance.
\subsection{Experiment Results}
\subsubsection{Single-class Door Detection}
Single-class door detection serves as the foundation of our approach, owing to the availability of existing single-class door datasets. Table~\ref{tab:singledoor} reports the mAP and mAP@50 of Co-DETR on the combined test set from \cite{doorobjectdetection, floorplans500}, after being trained on their combined training set. As shown in Table~\ref{tab:singledoor}, the model achieves a reasonably high mAP@50 of 0.793, indicating that it can reliably localize door instances with moderate IoU thresholds. However, the overall mAP score of 0.403 suggests that precise localization remains challenging under stricter thresholds.

\begin{table*}[t]
    \centering
    \caption{Single-class door detection results with Co-DETR on the combined dataset.}
    \begin{tabular}{c|c|c}
        \toprule
        \textbf{Method} & \textbf{mAP} & \textbf{mAP@50} \\
        \midrule
        Co-DETR (single-class) & 0.403 & 0.793 \\
        \bottomrule
    \end{tabular}
    \label{tab:singledoor}
\end{table*}
\subsubsection{Multi-class Door Detection}
Multi-class door detection is evaluated on our \textbf{DoorDet} dataset. We benchmark four state-of-the-art object detection methods: Co-DETR \cite{zong2023detrs}, InternImage-H \cite{wang2023internimage}, Focal-Stable-DINO \cite{ren2023strong}, and EVA \cite{fang2023eva}, which currently rank among the top-performing models on the COCO benchmark \cite{lin2014microsoft}.

\begin{figure*}[t]
    \centering
    \includegraphics[width=0.3\linewidth]{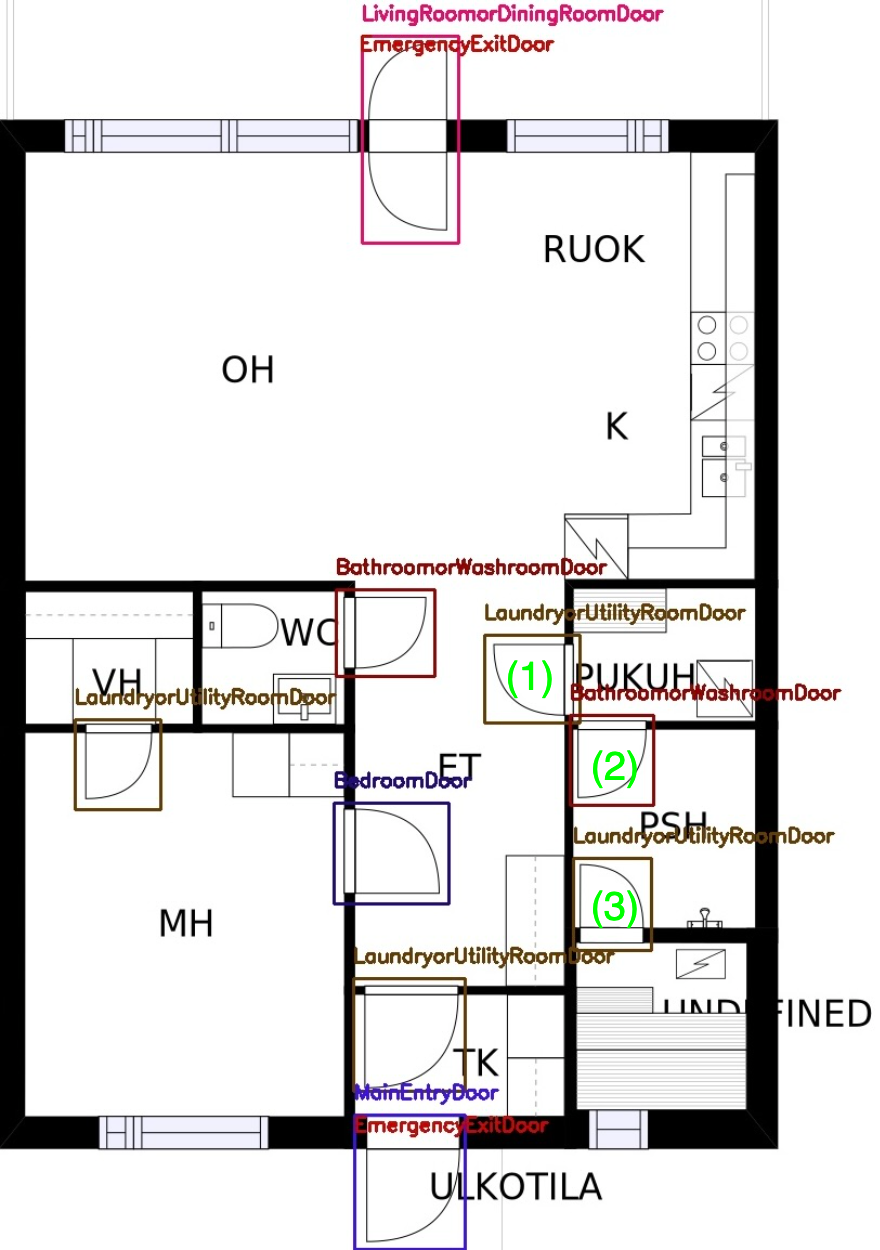}
    \caption{An example illustrating the challenges of multi-class door detection is shown in boxes (1), (2), and (3). In case (1), the door provides the only access to the room labeled `PUKUH', so its type is inferred based on the category of `PUKUH'. Similarly, the type of door in (2) is determined by its connection to `PSH', while in (3), the door type is inferred from the interior room it is part of. This demonstrates that the classification of a door may depend on the room it overlaps with in some cases, while in others, it relies on the adjacent room. Such variability in contextual cues makes the task inherently complex. However, the categories of all three doors were correctly predicted by GPT-4.1.
    \label{fig:complex}}
\end{figure*}

Table~\ref{tab:multipledoor} presents the results of different methods, reporting both mAP and mAP@50 for each class. These results establish a baseline for future research on our \textbf{DoorDet} dataset and highlight the challenging nature of the multi-class door detection task. Notably, the performance rankings are consistent with those observed on the COCO benchmark, although the performance gaps among methods are more pronounced. Figure~\ref{fig:complex} further illustrates our dataset's inherent complexity for multi-class door detection.

\begin{table*}
\small
\setlength{\tabcolsep}{2.5pt}
\centering
\caption{Benchmarking results of different multi-class door detection methods on the \textbf{DoorDet} dataset. `CD' is short for `Co-DETR'. `IIH' is short for `InternImage-H'. `FSD' is short for `Focal-Stable-DINO'.}
\begin{tabular}{l|c|c|c|c|c|c|c|c}
\toprule
\multirow{2}{*}{\textbf{Category}} & \multicolumn{4}{c|}{\textbf{mAP}} & \multicolumn{4}{c}{\textbf{mAP@50}} \\
\cmidrule(lr){2-5} \cmidrule(lr){6-9} & CD & IIH & FSD & EVA & CD & IIH & FSD & EVA \\
\midrule
Main entry door & \textbf{0.907} & 0.902 & 0.887 & 0.882 & \textbf{0.996} & 0.991 & 0.989 & 0.975 \\
Bedroom door & \textbf{0.891} & 0.883 & 0.861 & 0.851 & \textbf{0.983} & 0.979 & 0.966 & 0.955 \\
Bathroom or washroom door & \textbf{0.887} & 0.879 & 0.868 & 0.848 & \textbf{0.982} & 0.979 & 0.979 & 0.966 \\
Kitchen door & \textbf{0.852} & 0.808 & 0.814 & 0.763 & \textbf{0.924} & 0.890 & 0.899 & 0.844 \\
Living room or dining room door & \textbf{0.805} & 0.797 & 0.755 & 0.774 & \textbf{0.919} & 0.916 & 0.880 & 0.898 \\
Laundry or utility room door & \textbf{0.853} & 0.841 & 0.806 & 0.813 & \textbf{0.966} & 0.961 & 0.943 & 0.941 \\
Garage door & \textbf{0.806} & 0.729 & 0.705 & 0.781 & 0.885 & 0.809 & 0.864 & \textbf{0.902} \\
Balcony or terrace door & 0.841 & 0.834 & \textbf{0.855} & 0.814 & \textbf{0.956} & 0.948 & 0.954 & 0.927 \\
Emergency exit door & \textbf{0.872} & 0.870 & 0.858 & 0.551 & 0.988 & \textbf{0.990} & 0.987 & 0.664 \\
Study room door & 0.598 & \textbf{0.599} & 0.409 & 0.458 & 0.665 & \textbf{0.666} & 0.501 & 0.500 \\
\midrule
All & \textbf{0.831} & 0.814 & 0.782 & 0.753 & \textbf{0.926} & 0.913 & 0.896 & 0.857 \\
\bottomrule
\end{tabular}
\label{tab:multipledoor}
\end{table*}

\subsection{Ablation Studies}
\textit{\textbf{LLM:}} Without incorporating the LLM, door type cannot be performed automatically, and we must rely entirely on manual annotation. In this context, the benefit of using an LLM lies in significantly reducing annotation time by automating the initial process. We consider two manual alternatives: (1) labeling all door instances from scratch, and (2) performing door detection first, followed by human-in-the-loop refinement to assign door categories.

To validate the benefits of employing an LLM, we further provide statistics on the time required for refinement and compare it with the time needed for the above two alternatives. We randomly selected six examples and annotated them using labelImg\footnote{\url{https://github.com/tzutalin/labelImg}} for each of the three methods. Prior to annotation, we prepared the category list in the tool so that labels could be easily selected from the predefined options. Additionally, since the bounding boxes generated by our proposed pipeline do not perfectly align with object boundaries, we did not enforce absolute alignment when drawing new bounding boxes to ensure a fair comparison. Notably, since annotators may remember labels when annotating the same samples multiple times, we adopt a counterbalancing strategy to mitigate such effects. Specifically, we divide the six samples into three subsets, each containing two samples with comparable labeling difficulty. The three subsets are first annotated separately, each using one of the three annotation methods. For the first subset, annotation is performed from scratch, including manually drawing bounding boxes. For the second subset, we apply our proposed pipeline incorporating the LLM. For the third, we first convert all labels to a single category (``door'') and then refine them to their specific door types. After this initial phase, we proceed to annotate the remaining examples using each method. In total, all six samples are labeled using each approach. We record and average the time required for each method. This counterbalanced design minimizes the influence of memory or familiarity with the samples, ensuring a fair comparison of annotation efficiency.

Table~\ref{tab:labeltime} compares the manual annotation time across three methods: labeling from scratch, door detection with human-in-the-loop refinement (without LLM), and our proposed pipeline incorporating an LLM. From Table~\ref{tab:labeltime}, we observe that our method significantly reduces annotation time and associated effort compared to labeling from scratch. Moreover, integrating the LLM further improves efficiency over the refinement-only approach. Notably, even without the LLM, combining door detection with human feedback still outperforms manual annotation in terms of speed. Owning to the remarkable capabilities of modern LLMs, our proposed dataset generation pipeline can be readily extended to other tasks where annotation time is limited but strong performance is still desired. For instance, if we have a large collection of raw samples such as those in FloorPlanCAD \cite{fan2021floorplancad}, our proposed pipeline can rapidly generate a domain-specific object detection dataset, enabling efficient construction within a limited time budget.

\begin{table*}[t]
    \centering
    \caption{Per-Image Annotation Time: Manual Labeling vs. Human-in-the-Loop with and without LLM.}
    \begin{tabular}{l|c}
        \toprule
        \textbf{Annotation Method} & \textbf{Average Time (seconds)} \\
        \midrule
        Manual labeling from scratch & 97.5 \\
        Door Detection + Human-in-the-Loop & 70.0 \\
        Door Detection + LLM + Human-in-the-Loop & \textbf{54.5} \\
        \bottomrule
    \end{tabular}
    \label{tab:labeltime}
\end{table*}

\begin{figure*}[t]
    \centering
    \includegraphics[width=\linewidth]{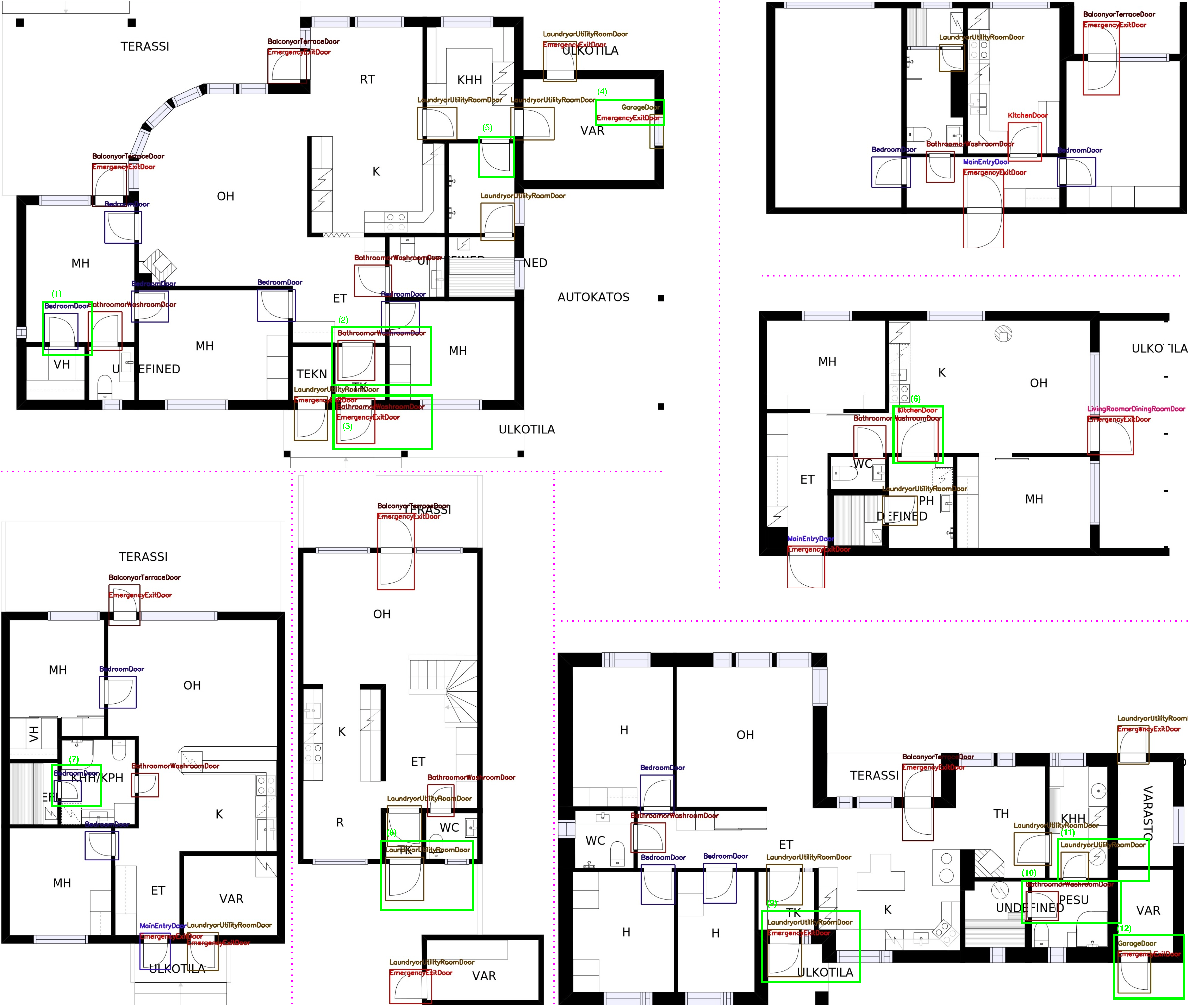}
    \caption{Examples of door type prediction with GPT-4.1. Boxes except green are predicted by GPT-4.1. Green boxes are found by us that are wrongly predicted.
    \label{fig:doortype}}
\end{figure*}

Furthermore, to provide an intuitive understanding of the benefits brought by LLMs, we present several examples of model-generated results prior to human-in-the-loop refinement. Figure~\ref{fig:doortype} presents these results. As shown, GPT-4.1 accurately predicts the door types in most cases.

\textit{\textbf{Human-in-the-Loop:}} The human-in-the-loop mechanism offers two key benefits. First, it substantially reduces annotation time, as demonstrated in Table~\ref{tab:labeltime}. Second, by incorporating human corrections into the pipeline, it has the potential to enhance the overall object detection performance.

To validate this, we conduct additional experiments on Co-DETR using the coarse dataset prior to human-in-the-loop refinement and compare the detection performance before and after the human-in-the-loop process. Table~\ref{tab:human-in-the-loop} presents the comparative results. The performance improves significantly after refinement, attributable to the more accurate and consistent annotations introduced through human feedback.

\begin{table*}[t]
    \centering
    \caption{Effect of human-in-the-loop refinement on Co-DETR for multi-class door detection.}
    \begin{tabular}{c|c|c}
        \toprule
        \textbf{Method} & \textbf{mAP} & \textbf{mAP@50} \\
        \midrule
        Co-DETR (w/o refinement) & 0.578 & 0.650 \\
        Co-DETR (w/ refinement) & \textbf{0.831} & \textbf{0.926} \\
        \bottomrule
    \end{tabular}
    \label{tab:human-in-the-loop}
\end{table*}
\subsection{Correlation Between Task Difficulty and Human Feedback}
Table~\ref{tab:perdoor} presents the per-category performance of the Co-DETR model on the \textbf{DoorDet} dataset, both \textbf{with} and \textbf{without} human-in-the-loop refinement. Additionally, it reports the task difficulty $D(t)$ and performance gain $\Delta(t)$, as defined in Equations~\ref{eq:difficulty} and~\ref{eq:gain}, respectively. As shown in Table~\ref{tab:multipledoor}, we observe that $D(t)$ and $\Delta(t)$ exhibit a positive correlation in most cases, indicating that categories with higher task difficulty generally benefit more from human-in-the-loop refinement.

\begin{table*}
\setlength{\tabcolsep}{2pt}
\centering
\caption{Per-category performance of Co-DETR on the \textbf{DoorDet} dataset, comparing results \textbf{with} and \textbf{without} human-in-the-loop refinement.}
\begin{tabular}{l|c|c|c|c|c|c|c|c}
\toprule
\multirow{2}{*}{\textbf{Category}} & \multicolumn{2}{c|}{\textbf{mAP}} & \multirow{2}{*}{\textbf{$D(t)$}} & \multirow{2}{*}{\textbf{$\Delta(t)$}} & \multicolumn{2}{c|}{\textbf{mAP@50}} & \multirow{2}{*}{\textbf{$D(t)$}} & \multirow{2}{*}{\textbf{$\Delta(t)$}} \\
\cmidrule(lr){2-3} \cmidrule(lr){6-7}& w/o & w/ & & & w/o & w/ & & \\
\midrule
Main entry door & 0.769 & 0.907 & 0.231 & 0.138 & 0.848 & 0.996 & 0.152 & 0.148 \\
Bedroom door & 0.813 & 0.891 & 0.187 & 0.078 & 0.894 & 0.983 & 0.106 & 0.089 \\
Bathroom or washroom door & 0.741 & 0.887 & 0.259 & 0.146 & 0.822 & 0.982 & 0.178 & 0.160 \\
Kitchen door & 0.734 & 0.852 & 0.266 & 0.118 & 0.796 & 0.924 & 0.204 & 0.128 \\
Living room or dining room door & 0.276 & 0.805 & 0.724 & 0.529 & 0.319 & 0.919 & 0.681 & 0.600 \\
Laundry or utility room door & 0.633 & 0.853 & 0.367 & 0.220 & 0.723 & 0.966 & 0.277 & 0.243 \\
Garage door & 0.328 & 0.806 & 0.672 & 0.478 & 0.370 & 0.885 & 0.630 & 0.515 \\
Balcony or terrace door & 0.688 & 0.841 & 0.312 & 0.153 & 0.801 & 0.956 & 0.199 & 0.155 \\
Emergency exit door & 0.802 & 0.872 & 0.198 & 0.070 & 0.928 & 0.988 & 0.072 & 0.060 \\
Study room door & 0.000 & 0.598 & 1.000 & 0.598 & 0.000 & 0.665 & 1.000 & 0.665 \\
\midrule
All & 0.578 & 0.831 & 0.422 & 0.253 & 0.650 & 0.926 & 0.350 & 0.276 \\
\bottomrule
\end{tabular}
\label{tab:perdoor}
\end{table*}
\subsection{Illustrations}
\begin{figure*}[t]
  \centering
  \begin{subfigure}[b]{0.24\textwidth}
    \includegraphics[width=\linewidth]{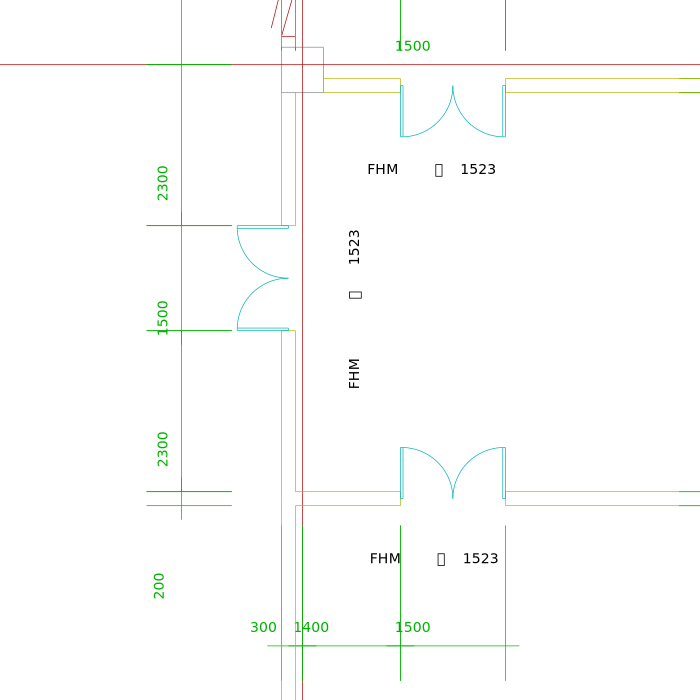}
  \end{subfigure}
  \hfill
  \begin{subfigure}[b]{0.24\textwidth}
    \includegraphics[width=\linewidth]{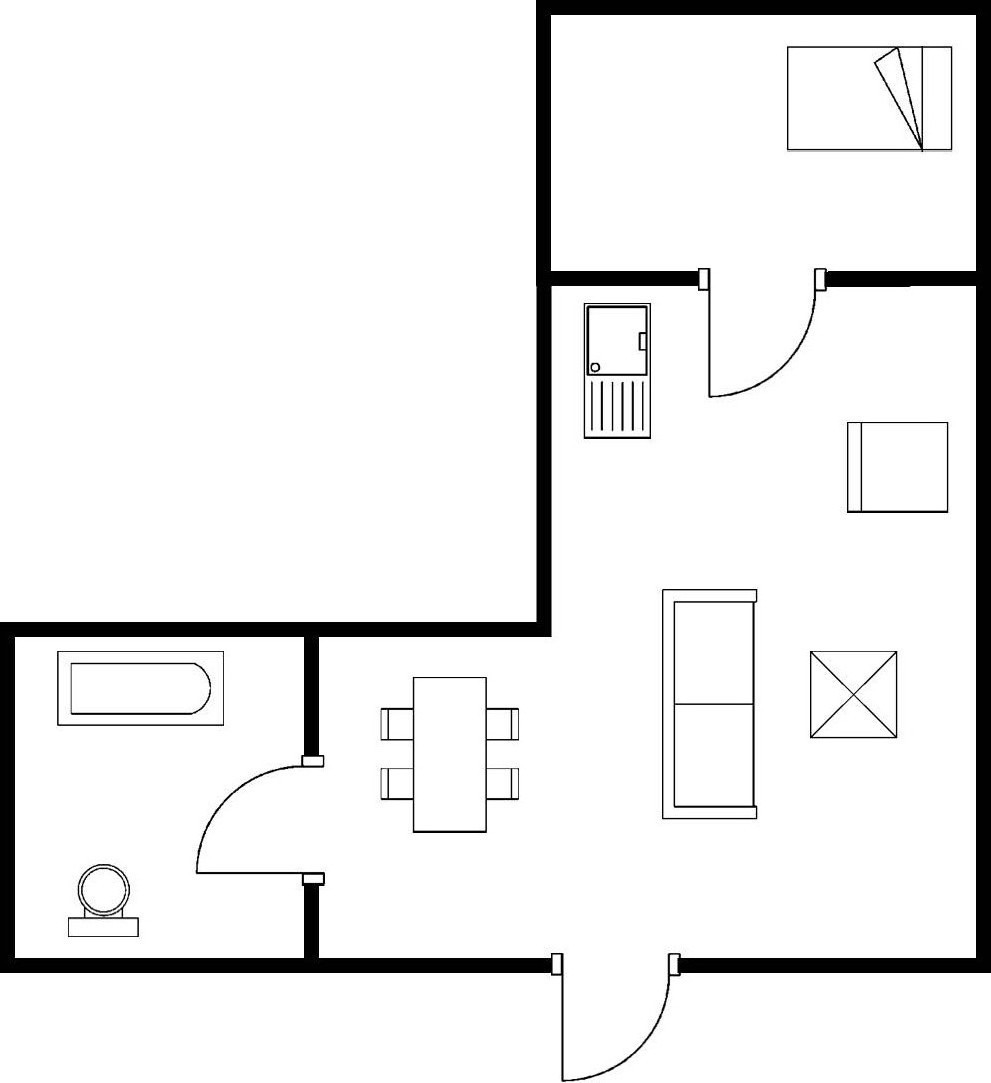}
  \end{subfigure}
  \hfill
  \begin{subfigure}[b]{0.24\textwidth}
    \includegraphics[width=\linewidth]{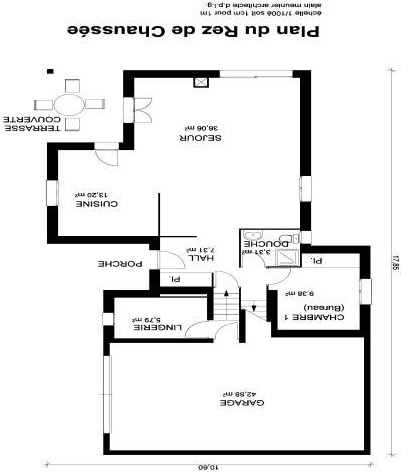}
  \end{subfigure}
  \hfill
  \begin{subfigure}[b]{0.24\textwidth}
    \includegraphics[width=\linewidth]{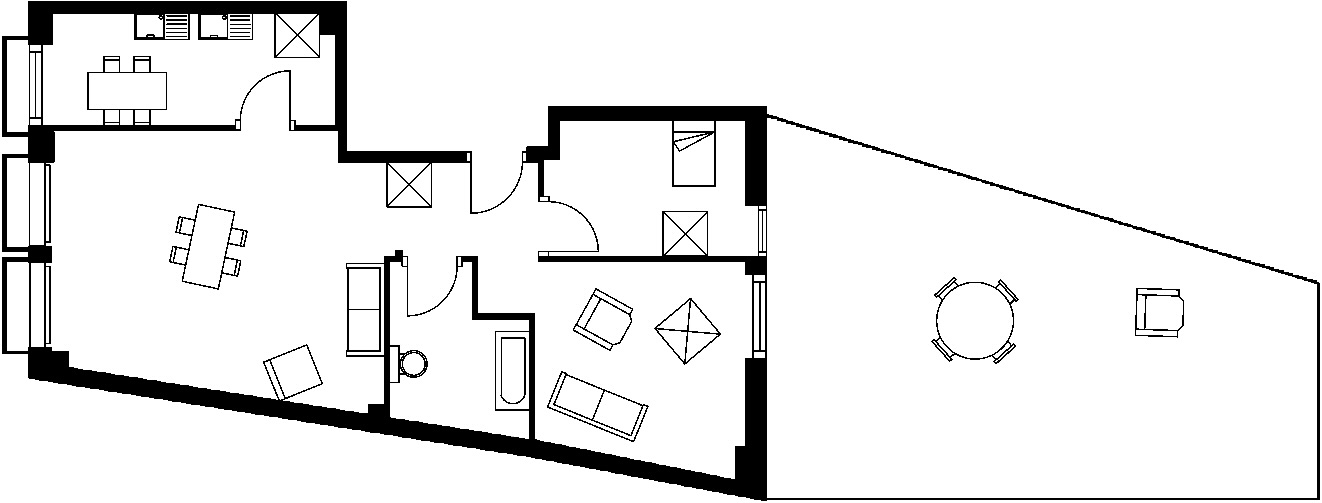}
  \end{subfigure}
  \vspace{0.5cm}
  \begin{subfigure}[b]{0.24\textwidth}
    \includegraphics[width=\linewidth]{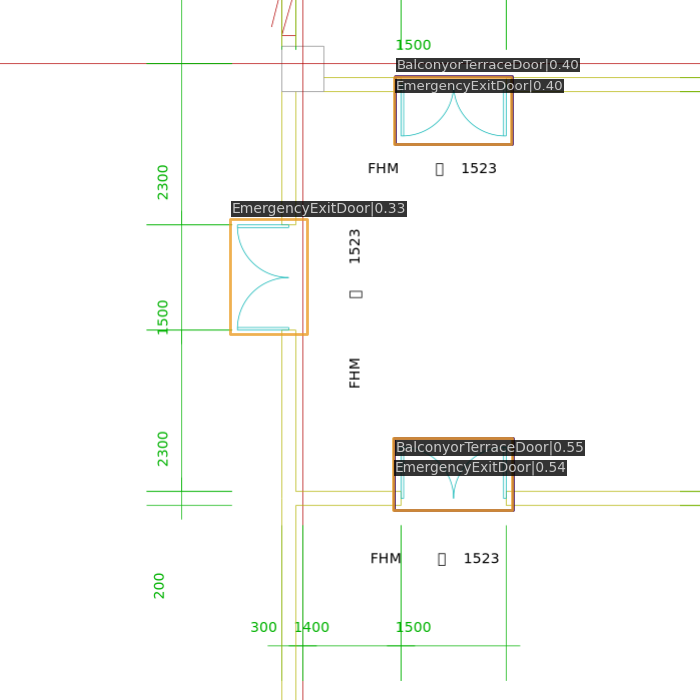}
    \caption{FloorPlanCAD \cite{fan2021floorplancad}}
  \end{subfigure}
  \hfill
  \begin{subfigure}[b]{0.24\textwidth}
    \includegraphics[width=\linewidth]{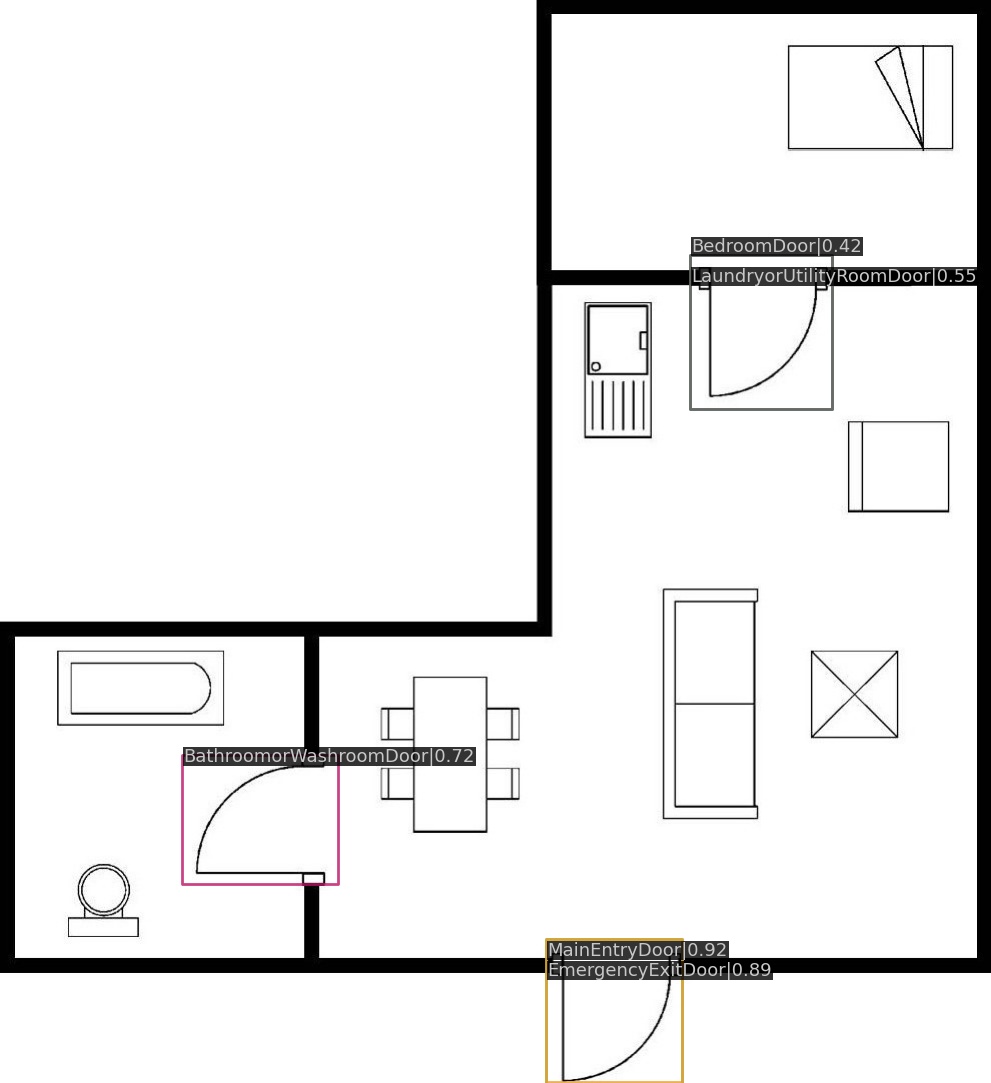}
    \caption{ROBIN \cite{sharma2017daniel}}
  \end{subfigure}
  \hfill
  \begin{subfigure}[b]{0.24\textwidth}
    \includegraphics[width=\linewidth]{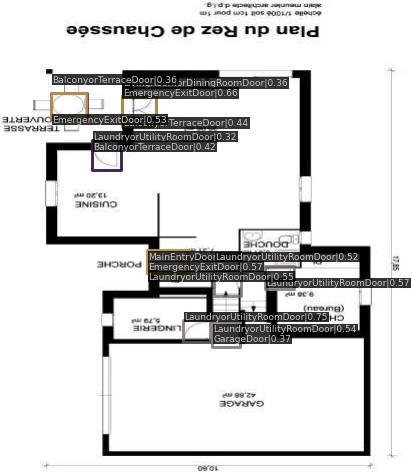}
    \caption{CVC-FP \cite{de2015cvc}}
  \end{subfigure}
  \hfill
  \begin{subfigure}[b]{0.24\textwidth}
    \includegraphics[width=\linewidth]{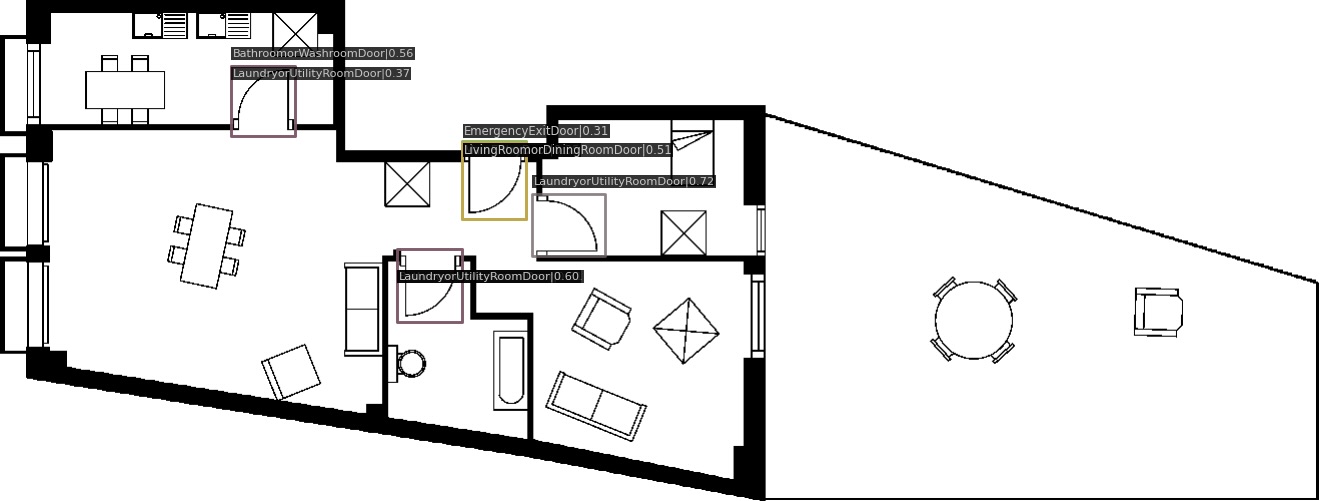}
    \caption{SESYD \cite{delalandre2010generation}}
  \end{subfigure}

  \caption{We conduct inference with the optimized Co-DETR model trained on our proposed \textbf{DoorDet} dataset using several datasets from other domains.}
  \label{fig:domainadaption}
\end{figure*}

Figure \ref{fig:domainadaption} illustrates several examples from other domains tested using the CO-DETR model optimized on our proposed \textbf{DoorDet} dataset. As shown, our dataset enables the trained object detection model to generalize to other datasets to some extent.
\section{Limitations and Future Work}
One limitation of our work lies in the fact that all bounding boxes in our dataset are algorithmically generated, and therefore only approximate rather than perfectly accurate. Nevertheless, we make efforts to refine obvious errors, particularly when bounding boxes significantly omit parts of the true door or noticeably deviate from object boundaries, manually correcting such substantial inaccuracies to improve overall quality. As shown in Figure~\ref{fig:doortype}, the bounding boxes generally provide sufficient coverage of the objects, typically extending only slightly beyond their true boundaries. To further alleviate the limitation, given the approximate nature of the annotations, we adopt mAP@50 as a more tolerant evaluation metric to accommodate minor localization inaccuracies, thereby providing a practical basis for benchmarking future methods.

As part of our future work, a promising direction is to explore effective methods for improving object detection performance on our dataset, particularly by leveraging techniques that address class imbalance. Another possible direction is to develop domain adaptation techniques to extend the use of our dataset to other domains.
\section{Conclusion}\label{sec6}
In this paper, we propose a novel approach to construct a highly useful multi-class door detection dataset in the field of floor plan drawings with minimal effort. We first detect all doors as a unified category using a state-of-the-art object detector, and then leverage the reasoning capabilities of an LLM to estimate door types based on their visual features and contextual cues. We validate our approach through extensive experiments, which demonstrate the effectiveness of our proposed method and the usefulness of the \textbf{DoorDet} dataset in terms of both accuracy and annotation efficiency.


\vspace{1em}

\noindent\textbf{Data availability} The dataset supporting the results of this study will be made publicly available upon acceptance of the manuscript. The data repository link and DOI will be provided at that time.

\section*{Declarations}
\textbf{Conflict of interest} The authors declare that they have no conflict of interest.


\bibliography{sn-bibliography}

\end{document}